\documentclass[10pt,twocolumn,letterpaper]{article}

\usepackage[pagenumbers]{cvpr} %

\definecolor{cvprblue}{rgb}{0.21,0.49,0.74}
\usepackage[pagebackref,breaklinks,colorlinks,allcolors=cvprblue]{hyperref}
\usepackage[title]{appendix}
\usepackage{amsmath}
\usepackage{amssymb}

\def\paperID{12486} %
\def\confName{CVPR}
\def\confYear{2025}

\title{CoCoNO: Attention \underline{Co}ntrast-and-\underline{Co}mplete for Initial \underline{No}ise Optimization in Text-to-Image Synthesis}

\author{
    \textit{Aravindan Sundaram}$^{1}$ \quad
    \textit{Ujjayan Pal}$^{1}$ \quad
    \textit{Abhimanyu Chauhan}$^{2}$ \quad\\
    \textit{Aishwarya Agarwal}$^{3}$ \quad
    \textit{Srikrishna Karanam}$^{3}$ \quad\\[2mm]
    \small $^{1}$Indian Institute of Technology Madras, Chennai, India\\
    \small $^{2}$Indian Institute of Technology Bombay, Mumbai, India\\
    \small $^{3}$Adobe Research, Bengaluru, India\\[2mm]
    \small \texttt{mm21b010@smail.iitm.ac.in} \quad
    \texttt{cs21b084@smail.iitm.ac.in} \quad
    \texttt{200040006@iitb.ac.in} \quad\\
    \small \texttt{aishagar@adobe.com} \quad
    \texttt{skaranam@adobe.com} 
}

\begin{document}
\twocolumn[{
\renewcommand\twocolumn[1][]{#1}%
\maketitle
\begin{center}
 \centering
 \captionsetup{type=figure}
 \includegraphics[width=\textwidth]{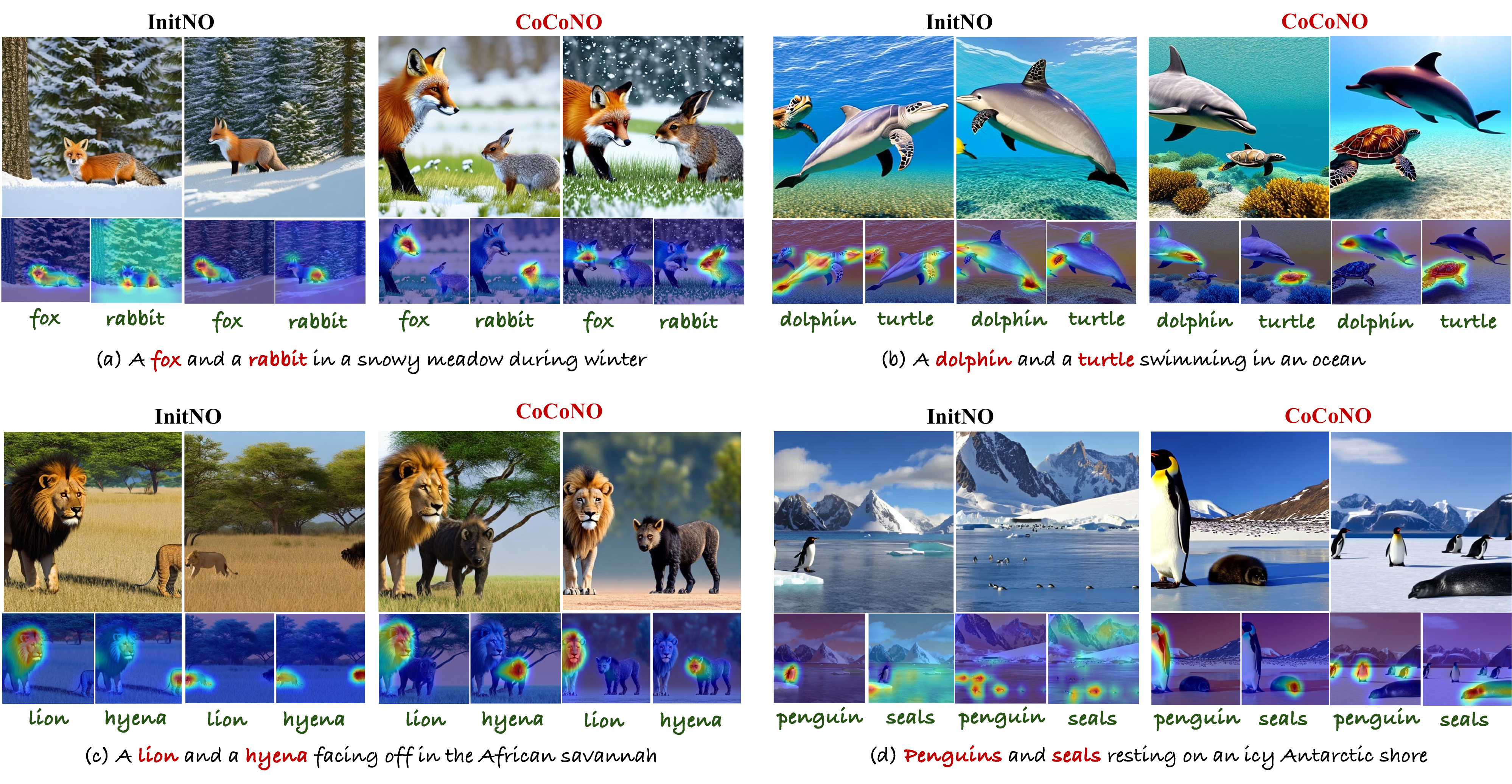}
 \vspace{-8pt}
 \caption{We propose CoCoNO, a new algorithm for determining the best initial latent noise which when denoised with text-to-image models can result in a more input-prompt-aligned image output. The current state-of-the-art method InitNO often misses out on key subjects (e.g., rabbit in (a)) or produces mixed-up images (e.g., dolphin fins and face that look like turtle in (b)). CoCoNO addresses these issues with two new loss functions, attention contrast and attention complete, which when used to optimize the initial noise can produce much more meaningful images (see results tagged ``CoCoNO").}
 \label{fig:teaser_qual}
\end{center}
}]
\begin{abstract}
Despite recent advancements in text-to-image models, achieving semantically accurate images in text-to-image diffusion models is a persistent challenge. While existing initial latent optimization methods have demonstrated impressive performance, we identify two key limitations: (a) \textbf{attention neglect}, where the synthesized image omits certain subjects from the input prompt because they do not have a designated segment in the self-attention map despite despite having a high-response cross-attention, and (b) \textbf{attention interference}, where the generated image has mixed-up properties of multiple subjects because of a conflicting overlap between cross- and self-attention maps of different subjects.

To address these limitations, we introduce \textbf{CoCoNO}, a new algorithm that optimizes the initial latent by leveraging the complementary information within self-attention and cross-attention maps. Our method introduces two new loss functions: the \textbf{attention contrast} loss, which minimizes undesirable overlap by ensuring each self-attention segment is exclusively linked to a specific subject's cross attention map, and the \textbf{attention complete} loss, which maximizes the activation within these segments to guarantee that each subject is fully and distinctly represented. Our approach operates within a noise optimization framework, avoiding the need to retrain base models. Through extensive experiments on multiple benchmarks, we demonstrate that CoCoNO significantly improves text-image alignment and outperforms the current state of the art.
\end{abstract}
    
\section{Introduction}
\label{sec:intro}

\begin{figure}[]
    \centering
    \includegraphics[width = 1\linewidth]{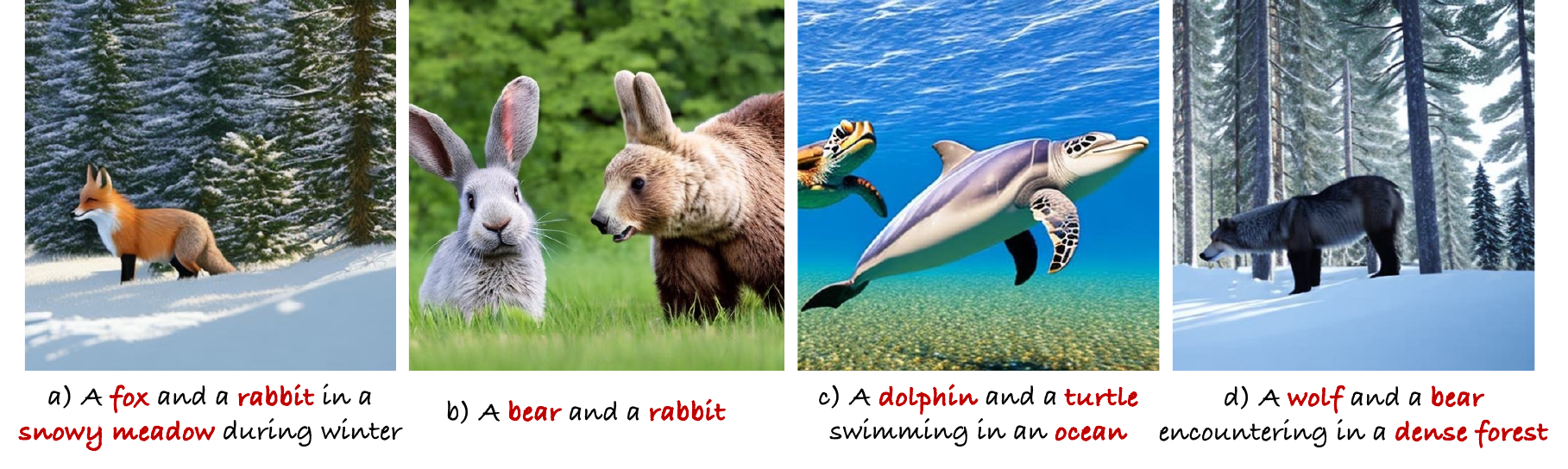}
    \caption{Images demonstrating subject mixing and neglect.}
    \label{fig:limnpre}
\end{figure}

While recent text-to-image models \cite{ramesh2022hierarchical, saharia2022photorealistic, rombach2022high} have enabled highly-realistic image synthesis given freeform text, one persistent challenge with them is achieving precise alignment between the generated images and the input text. There are multiple issues, including i) \textit{subject neglect} where the model omits generating certain subjects (e.g., missing \texttt{rabbit} and \texttt{bear} in Figure \ref{fig:limnpre} (a) $\&$ (d) respectively); ii) \textit{subject mixing} where the model generates subjects with mixed properties (e.g. the \texttt{bear} with \texttt{rabbit's ears} in \ref{fig:limnpre} (b)).

Recent work that seeks to alleviate these issues can broadly be categorized into test-time latent optimization \cite{agarwal2023star, chefer2023attend}, initial noise optimization \cite{guo2024initno, eyring2024reno}, and prompt optimization \cite{manas2024improving, witteveen2022investigating, hertz2022prompt, fan2024prompt, mo2024dynamic, hao2024optimizing}, apart from several training-based approaches (involving model parameter update) \cite{mo2024dynamic, hao2024optimizing}. For example, CONFORM \cite{meral2024conform}, A-STAR \cite{agarwal2023star} and Attend-and-excite \cite{chefer2023attend} optimize the cross-attention maps during denoising at test time to ensure subject presence, and InitNO \cite{guo2024initno} optimizes the initial latent noise to ensure a good starting point for the subsequent denoising process. 
Test-time latent optimization approaches present several limitations, including the restricted availability of denoising timesteps necessary for the convergence of losses and the potential for out-of-distribution shifts \cite{chefer2023attend} when iteratively refining the latent codes. Prompt optimization methods \cite{manas2024improving} start from a user prompt and iteratively generate revised prompts with the goal of maximizing a consistency score between the generated images and prompt. However, prompt updates in these methods are not a direct function of losses computed over latents, thereby not ensuring/guaranteeing the desired next step. Training-based methods are not very flexible since they involve model retraining, and this is inflexible to do for various reasons, e.g., compute availability. Consequently, our focus in this work is to optimize the initial latent so as to obtain a good starting point which when denoised can address the issues above.

\begin{figure*}[]
    \centering
    \includegraphics[width = 0.93\linewidth]{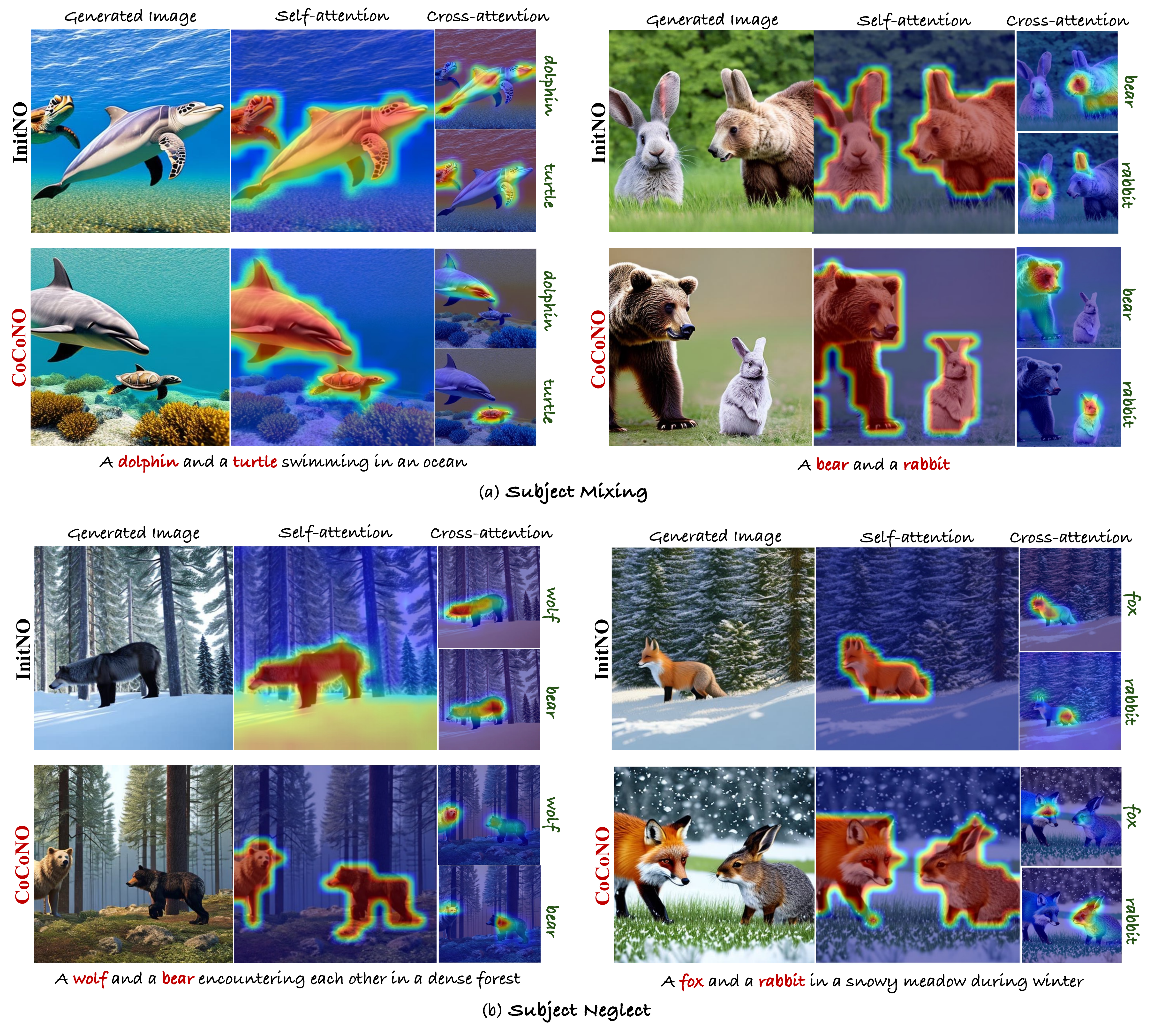}
    \caption{Our proposed CoCoNO alleviates subject neglect and mixing by ensuring one high-response self-attention segment for each subject (e.g., wolf and bear in second row), while minimizing interference between the cross-attention map for a subject (e.g., turtle in first row) with the self-attention segment of other subjects (dolphin here).}
    \label{fig:limnDetailsMaps}
\end{figure*}

While initial noise optimization methods such as InitNO \cite{guo2024initno} have shown impressive performance, they fail to correctly identify the underlying causes of subject mixing and subject neglect, thereby leading to suboptimal improvements. 
Consider the example in Figure~\ref{fig:limnDetailsMaps} (a) where we show generated images and the corresponding cross- and self-attention maps using InitNO \cite{guo2024initno} and our proposed method, CoCoNO. InitNO attempts to alleviate subject mixing by minimizing the overlap between the self-attention maps of a pair of subject tokens. However, ensuring non-overlapping self-attention is insufficient to tackle this issue, e.g., the \texttt{turtle-like} texture visible on \texttt{dolphin's fins} and \texttt{mouth}, even though turtle and dolphin attention maps have no overlap. This is due to both \texttt{turtle} and \texttt{dolphin} attempting to attend to a common object/segment, as can be noted from the self-attention map (notice parts of \texttt{turtle's} cross-attention maps attend to the segment for \texttt{dolphin} in the self-attention map/generated image, and this is exactly same as the regions in \texttt{dolphin} having \texttt{turtle-like} properties). A similar phenomenon can be noticed in the second example in Figure~\ref{fig:limnDetailsMaps} (a) (\texttt{bear} has \texttt{rabbit}'s ears, cross-attention maps have no overlap but \texttt{rabbit}'s cross-attention map attends to \texttt{bear}'s ears in self-attention). Across both these examples, we notice highly-activated regions in cross-attention maps attending to overlapping regions in self-attention maps, i.e., cross-attention map of token A attending to token B's region in the self-attention map. We call this issue \textbf{attention interference}.

Next, to address subject neglect, InitNO ensures the presence of high-activation regions in each subject token's cross-attention maps. But from the first column in Figure~\ref{fig:limnDetailsMaps} (b), this is insufficient, e.g., both \texttt{wolf} and \texttt{bear} have highly activated cross-attention maps, but that does not help as there is only one segment in the self-attention map and both cross-attention regions attend to this segment. Another example is in the second column where, again, since the self-attention map has only one segment for InitNO, cross-attention regions for both \texttt{rabbit} and \texttt{fox} attending to the same segment. Across both these examples, we notice the primary issue of missing segments in the self-attention maps (both examples had two subject tokens but only one contiguous self-attention segment), which leads to cross-attention maps of all tokens attending to this same self-attention region. We call this issue \textbf{attention neglect}.

To tackle the aforementioned issues, we propose \textbf{CoCoNO}, a novel algorithm for optimizing the initial noise which we denoised with a text-to-image model will give the desired image. Unlike InitNO \cite{guo2024initno}, our key idea is to leverage self- and cross-attention maps jointly in designing new loss functions since they have complementary pieces of information. To this end, \textbf{CoCoNO} first creates a mapping between each token's cross-attention map and the corresponding segments obtained from the self-attention map, leading to an assignment of each self-attention segment to a particular token. %
Given this assignment, our proposed \textbf{attention complete loss} maximizes the cross-attention activation of each subject within its assigned self-attention segment. Our key insight here is that each subject token should necessarily have a designated high-response segment in the self-attention map, thereby ensuring the presence of all subjects. Next, our proposed \textbf{attention contrast loss} minimizes the overlap between the cross-attention map of a subject and the self-attention segments of other subjects. This reduces the inter-subject confusion in the attention space, which we noted was a primary reason for subject mixing in Figure~\ref{fig:limnDetailsMaps} previously. We show some results with CoCoNO in Figure~\ref{fig:limnDetailsMaps}. Our method leads to a segment each for \texttt{dolphin} and \texttt{turtle} in the first example and cross-attention maps for dolphin and turtle only attend to the corresponding self-attention segments with no intermixing, leading to improved generation when compared to InitNO. This is because InitNO only ensures high-response and non-overlapping attention maps, which is insufficient to alleviate subject neglect and subject mixing as discussed previously.

Our key contributions are summarized below:
\begin{itemize}
    \item We identify \textbf{attention neglect} and \textbf{attention interference} as two key limitations of the current state-of-the-art initial latent optimization methods for text-to-image synthesis, leading to issues such as subject neglect and subject mixing (see Figure~\ref{fig:teaser_qual}).
    \item We propose two new loss functions to alleviate the above issues: the \textbf{attention complete loss} ensures each subject in the input prompt has a designated self-attention segment, and the \textbf{attention contrast loss} ensures there is no overlap between cross-attention maps of one subject with self-attention segments of other subjects.
    \item We conduct extensive experiments on standard benchmark datasets using standard evaluation metrics and show our method significantly outperforms the current state of the art.
\end{itemize}.

\section{Related Work}
\label{sec:relatedWorks}
Generative image synthesis has seen many breakthroughs in the last decade or so, starting with variational autoencoders \cite{kingma2013auto,huang2018introvae}, generative adversarial networks \cite{isola2017image, karras2019style, park2019semantic, zhu2017unpaired, zhu2017toward,tao2022df, xu2018attngan, zhang2021cross, zhu2019dm, ye2021improving}, and most recently diffusion models \cite{hao2024optimizing, xu2018attngan, zhu2019dm, tao2022df,ho2022classifier,rombach2022high}. 

Despite their dramatic success, getting images that align well with the input prompt is still a challenge, with a wide variety of methods attempting to address this issue, including prompt engineering \cite{manas2024improving, witteveen2022investigating, hertz2022prompt, fan2024prompt, mo2024dynamic, hao2024optimizing} and finetuning/reinforcement learning strategies \cite{mo2024dynamic, hao2024optimizing}. Some other methods that seek to address this issue include inference-time latent update methods such as Chefer et al. \cite{chefer2023attend} where cross-attention maps were manipulated by maximizing the activations of the most neglected concepts and Agarwal et al. \cite{agarwal2023star} where losses based on cross-attention overlap were used to update latents during each denoising step. However, one issue with these methods is they tend to produce out-of-distribution latents (because of latent updates during denoising), leading to misaligned images.

While initial latent optimization techniques \cite{manas2024improving,guo2024initno} handle these drawbacks, as discussed in Section 1, they very often lead to certain subjects being omitted or images having mixed-up properties of multiple subjects. We address these issues by identifying the underlying reasons and proposing two new loss functions that utilize both self- and cross-attention maps jointly in their design.

\section{Methodology}
\label{sec:method}

\subsection{Preliminaries}
We first briefly review latent diffusion models (LDMs) and the associated cross-/self-attention computation. LDMs comprise a latent encoder-decoder pair and a denoising diffusion probabilistic model (DDPM). In Rombach et al. \cite{rombach2022high}, the encoder-decoder pair is a standard variational autoencoder \cite{kingma2013auto, van2017neural} where an image $I\in R^{(w\times H\times3)}$ is mapped to a latent representation $z=E(I)\in R^{(h\ \times w\ \times c)}$ using E. A decoder D is trained to reconstruct $I\approx D\left(z\right)$. The DDPM then operates in the $z-$space in a series of denoising steps. In each step t, given $z_t$, the DDPM is trained to produce a denoised version $z_{t-1}$. In text-to-image LDMs, this DDPM is usually conditioned using text embeddings computed with a text encoder L (e.g., CLIP \cite{radford2021learning}, T5 \cite{raffel2020exploring}). Given input prompt p’s representation $L\left(p\right)$, the DDPM $\epsilon_\theta$ is trained to minimize:
\begin{equation}
    E_{(z~E(I),p,\epsilon~N(0,1),t)\ }\ [|(|\epsilon-\epsilon_\theta\ (z_t,L(p),t)|)|] 
\end{equation}
At test time, given $L\left(p\right)$, one can synthesize an image by repeatedly denoising the initial latent code $z_T \sim N (0, 1)$ in T steps. The denoised latent can then be decoded using D to get the image $I$. %

As shown in prior work \cite{guo2024initno, agarwal2023star}, text conditioning via cross-attention \cite{vaswani2017attention} layers results in a set of cross-attention maps $A_t\in R_{r\times r\times N}$ (r = 16 from Hertz et al. \cite{hertz2022prompt}) at each denoising step t for each of N tokens (tokenized using L’s tokenizer) in the input prompt p. For instance, see Figure 4 where the cross-attention maps of cow and buffalo are indeed highlighting the correct spatial location. In our implementation, we compute the overall cross-attention map by averaging the cross-attention maps across all layers and heads at a $16\times16$ resolution. Prior to averaging, we exclude the [sot] token's attention map and normalize all other tokens' maps. This results in an aggregated cross-attention map,\ $A^C\in\mathbb{R}^{16\times16\times n}$, containing $n$ spatial cross-attention maps for each subject token in the prompt. Similarly, the self-attention map (computed via PCA following \cite{tumanyan2023plug}) is aggregated across all layers, providing insight into how each pixel in the $16\times16$ map attends to every other pixel. These maps can be denoted as $A^S\in\mathbb{R}^{16\times16\times256}$. 
\subsection{CoCoNO: Attention \underline{Co}ntrast-and-\underline{Co}mplete for Initial \underline{No}ise Optimization}
As noted in Section 1, we identified two key issues with initial noise optimization methods such as InitNO \cite{guo2024initno} for text-to-image synthesis: attention interference and attention neglect. In attention interference, we observed that cross-attention maps of token A attend to self-attention regions designated for token B, leading to subject mixing (recall instances of \texttt{bear} with \texttt{rabbit's ears} and \texttt{dolphin} with \texttt{turtle's texture} in Figure~\ref{fig:limnDetailsMaps}(a)). In attention neglect, we observed that there are missing subject segments in the self-attention maps, leading to cross-attention maps of multiple subject tokens attending to the same self-attention segment (e.g., missing \texttt{bear} and \texttt{rabbit} in Figure~\ref{fig:limnDetailsMaps}(b)). To address these issues, we propose two new loss functions that use information from both cross- and self-attention maps jointly, unlike prior work \cite{guo2024initno} that operates on them independently. Before discussing the details of our new losses, we first explain our intuition.

\begin{figure}[h]
    \centering
    \includegraphics[width = \linewidth]{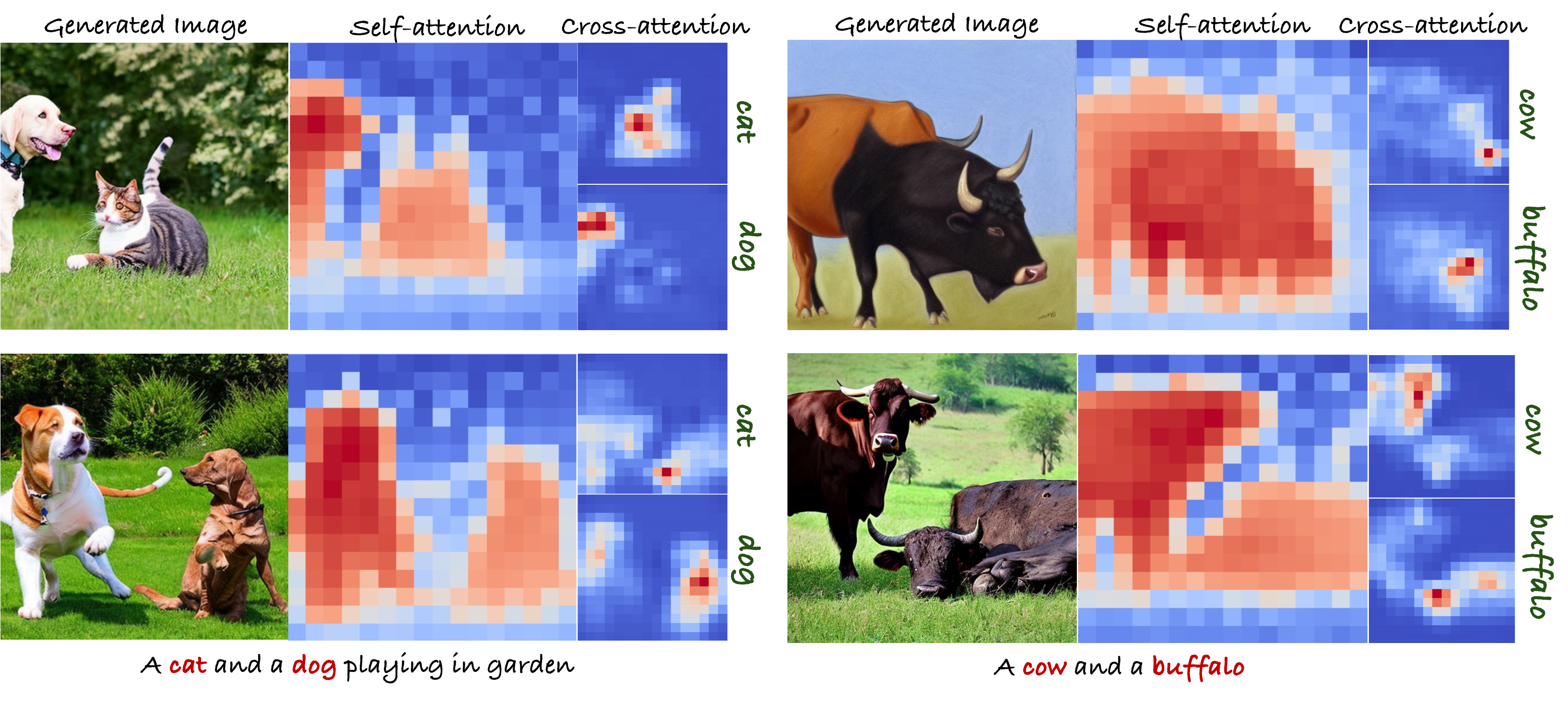}
    \caption{Intermediate one-step denoised attention maps.}
    \label{fig:attentionMapsVis}
\end{figure}

Our problem setup is similar to InitNO: to optimize the latent noise which when used as a starting point can give desired image outputs. To understand why this may work, consider Figure~\ref{fig:attentionMapsVis} where we show cross- and self-attention maps for a partially denoised latent (denoised for exactly one step). This figure shows these attention maps are reasonably indicative of the spatial location of the objects (e.g. see attention maps for the \texttt{cat} and \texttt{dog} in the first row). Additionally, these attention maps are sufficiently informative to deduce the potential \textbf{attention interference} and \textbf{neglect} issues discussed previously in Section 1. For instance, consider the example shown in first row/second column of Figure~\ref{fig:attentionMapsVis} where there is only one object present in the generated image instead of both cow and buffalo. As one can note from the attention maps, this is because the self-attention map here has only one segment, with the cross-attention maps for both \texttt{cow} and \texttt{buffalo} activated at different regions within the same segment. Similarly, in the second row/first column, even though the self-attention map has two segments here, and the cross-attention maps for both \texttt{cat} and \texttt{dog} are highly activated, we end up getting two dogs instead due to the cross-attention map of \texttt{dog} attending to both self-attention segments.

\begin{figure}[h]
    \centering
    \includegraphics[width = 1.01\linewidth]{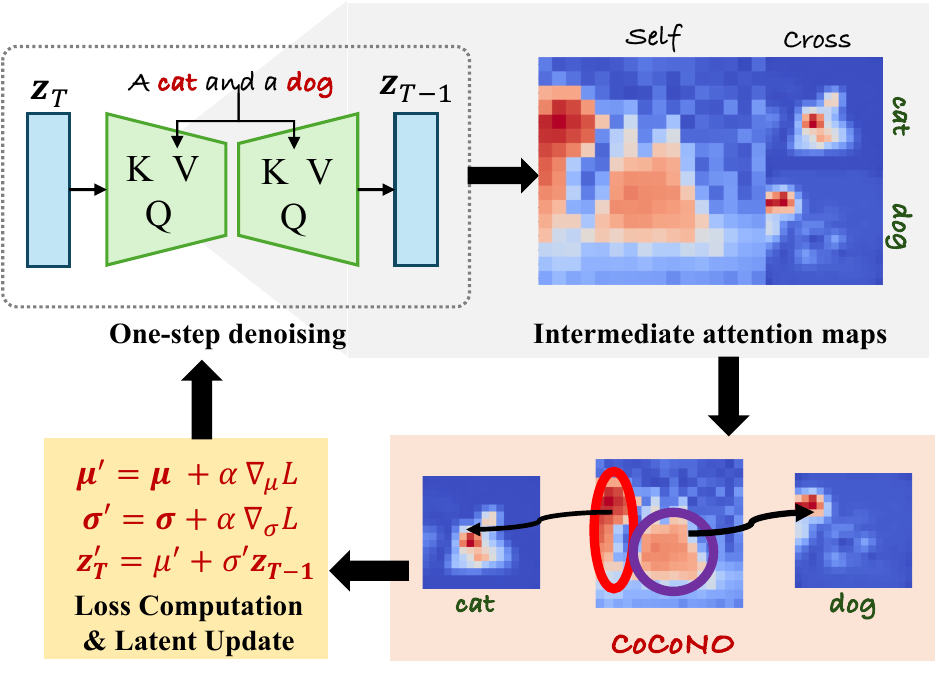}
    \caption{A visual illustration of proposed method.}
    \label{fig:pipelineFig}
\end{figure}

To address InitNO's limitations discussed above, we use these attention maps computed from such partially denoised latents and propose novel losses for latent optimization. The goal with our method (visually summarized in Figure~\ref{fig:pipelineFig}) is to ensure we obtain a starting latent which when denoised can ensure presence of all subjects in the prompt and minimize subject mixing as much as possible. Our intuition to achieve this is to ensure the self- and cross-attention maps hold exactly these properties, because that would mean an initial latent that has these properties. In particular, we want the following:

\begin{itemize}
    \item Cross-attention maps of any pair of subjects should not just have high-response regions but also attend to different segments/objects in the self-attention map/generated image. This also means self-attention maps should have the same number of unique segments as the subjects.
    \item There should be no interference between high-response regions in one subject's cross-attention map with self-attention regions corresponding to other subjects.
\end{itemize}

\subsection{Attention Contrast-and-Complete}

We next discuss how our intuition above translates to the specific losses that constitute CoCoNO (see Figure~\ref{fig:pipelineFig}). We first start by randomly sampling an initial latent code $\mathbf{z}_\mathbf{T}\sim\mathcal{N}(\mu,\sigma)$, where $(\mu,\sigma)$ are the parameters to be updated as part of the optimization process. We initialize them to be zero-mean and unit-covariance and then do one step of denoising to obtain $\mathbf{z}_\mathbf{T-1}$. Given $\mathbf{z}_\mathbf{T-1}$, we compute losses using our proposed method (discussed next) and update the $(\mu,\sigma)$ parameters as:
\begin{equation}
    \begin{split}
\mu^\prime &=\mu+\nabla_\mu \mathcal{L}  \\
\sigma^\prime &=\sigma+\nabla_\sigma \mathcal{L}
\end{split}
\end{equation}
Given the updated $\mu^\prime$ and $\sigma^\prime$ values, the latent code is then updated as:

\begin{equation}
\mathbf{z}_\mathbf{T}^\prime=\mu^\prime+\sigma^\prime\mathbf{z}_\mathbf{T-1}
\end{equation}

We then again do one-step denoising with this updated $\mathbf{z}_\mathbf{T}^\prime$, update the $(\mu,\sigma)$ parameters, and repeat this process until convergence, leading to the final optimized parameters $(\hat{\mu},\hat{\sigma})$ which will give us the desired starting latent. This starting latent is then used as the starting point to denoise and generate images with the input prompt. To clarify, the key difference between CoCoNO and InitNO \cite{guo2024initno} is how we formulate our loss $\mathcal{L}$- the rest of the parameter and latent update process is similar for consistency and ease of comparisons later. We next discuss how our loss $\mathcal{L}$ is computed. 

Given the latent code $z$ at any iteration of the procedure discussed above, we compute the self-attention maps and determine the first principal component. These maps are then sigmoid softened as $A=\sigma(\alpha(A-\beta))$ (where A is the self-attention map, $\sigma()$ is the sigmoid function, and $\alpha = 16$,  $\beta = 0.5$ are scalars) to exclude low response regions and increase the importance of high-response ones. For an input prompt comprising $n$ subjects, we separate these high-response regions to get $n$ distinct segments, one corresponding to each subject. Let $A^s \in \mathbb{R}^{r \times r \times n}$ denote the matrix representing these $n$ segments ($A^s_i$ refers to the $i^{th}$ segment). Note that it is possible for $A$ to have less than $n$ segments (e.g., $u<n$); in this case, we create $n-u$ zero-element matrices (of dimensions $r \times r$) so that there are $n$ segments in total. Intuitively, these zero-element matrices represent subjects that are omitted by the model (that we will correct with CoCoNO). We also compute the cross-attention maps corresponding to these $n$ subject tokens and denote them as $A^c \in \mathbb{R}^{r \times r \times n}$. Given the information above, we seek to assign each segment in the self-attention map to one distinct cross-attention map, resulting in a one-to-one mapping. We run an assignment optimization operation \cite{kuhn1955hungarian} to compute this mapping. Briefly, for a given cost function $C$, this operation determines an optimal permutation matrix $\hat{P}$ such that $Tr(PC)$ is maximized [$Tr(.)$ denotes the trace of matrix]. In our context, this translates to maximizing the intersection between the self-attention segments in $A^s$ and the cross-attention maps in $A^c$. We compute this matrix C as:
\begin{equation}
    C(i,j) = \sum_{k=1}^r \sum_{l=1}^r \left[ A_i^s \ast A_j^c \right]_{k,l}
\end{equation} 
where * denotes element-wise multiplication. $C(i, j)$ denotes the intersection between the $i^{th}$ self-attention segment $A_i^s$ and the $j^{th}$ cross-attention map $A_j^s$ where $i, j \in [1, n]$. The matrix $C$ then represents intersection values between each possible pair of self-attention segments and cross-attention maps. After optimization, every subject token will have a corresponding row in $\hat{P}$ that represents which self-attention segment it got mapped to (and that entry will have a 1, rest in the row will be zeros). For example, if the first row has values like $(0,0,1,0)$, it means the first subject got mapped to the third self-attention segment. This permutation matrix $\hat{P}$ informs our novel attention contrast and attention complete losses, discussed next.

First, as discussed above, we want no interference between the high-response regions in one subject's cross-attention map with segments of other subjects in the self-attention map. To realize this objective, we propose the \textbf{attention contrast} loss. In the permutation matrix $\hat{P}$, each zero-element entry corresponds to an undesired mapping between a subject's cross-attention map and some other subject's self-attention segment. We gather all such values from the $C$ matrix and simply minimize the resulting overall intersection value, thereby minimizing the interference noted above. This is implemented using the loss function below: 
\begin{equation}
L_\text{Acont} = \sum_{i,j,i \neq j} \left( \frac{\left[\hat{P} \otimes C \right]_{i,j}}{\sum_{k=1}^r \sum_{l=1}^r A_i^s(k,l)} \right)
\end{equation}

where $\otimes$ refers to the matrix multiplication operation and $\hat{P}\otimes C$ for $i\neq j$ (i.e., off-diagonal elements) gives us all the undesired intersection values.

Next, we want each subject's cross-attention map to have a designated and unique high-response segment in the self-attention map. This means we want the self-attention map to have $n$ complete segments each having a high overlap with its corresponding subject's cross-attention map (recall the missing segments above were set to zero matrices; we want these to be populated with actual non-zero values that represents the presence of a segment). We implement this by considering the diagonal elements in $\hat{P}\otimes C$, determining the element with the least/minimum overlap/intersection value, and maximizing this. This is achieved with our proposed \textbf{attention complete} loss:
\begin{equation}
\displaystyle L_\text{Acomp} = 1 - \min_{i} \left( \frac{\left[\hat{P} \otimes C \right]_{i,i}}{\sum_{k=1}^r \sum_{l=1}^r A_i^S(k,l)} \right)
\label{eqn:completion}
\end{equation}
Note that in a scenario where there are missing segments (e.g., missing \texttt{wolf} and \texttt{rabbit} in Figure~\ref{fig:limnDetailsMaps}), the minimum value will actually be zero, resulting in a high loss (loss value of 1). This loss will continue to be remain 1 as long as there is a missing segment, and minimizing this loss will give us the presence of this missed segment (i.e., the minimum overlap value will then be high). 

\begin{figure*}[]
    \centering
    \includegraphics[width = 0.95\linewidth]{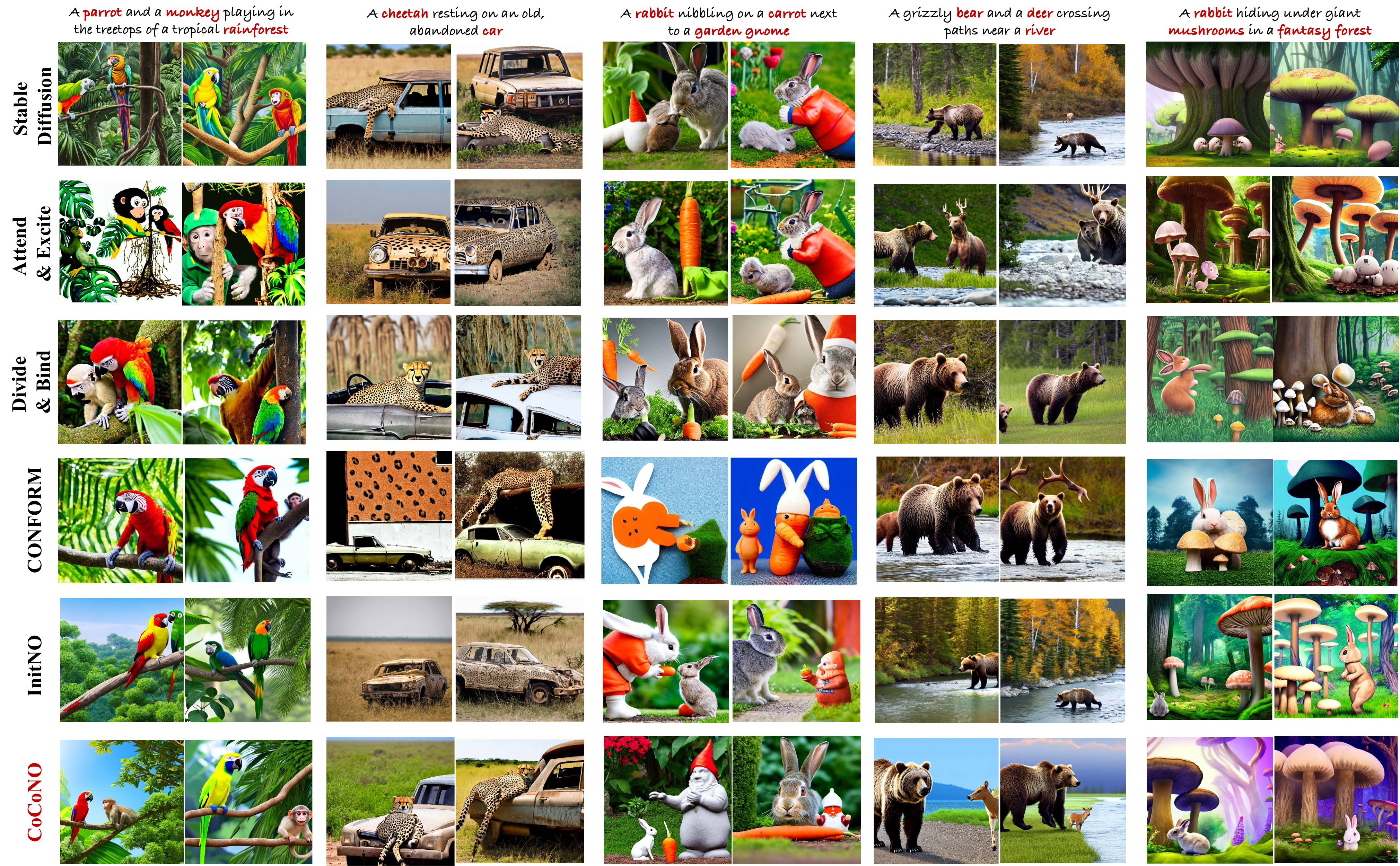}
    \caption{Qualitative comparisons of CoCoNO with recent state-of-the-art methods.}
    \label{fig:qualResults}
\end{figure*}

\begin{figure*}[]
    \centering
    \includegraphics[width = 0.95\linewidth]{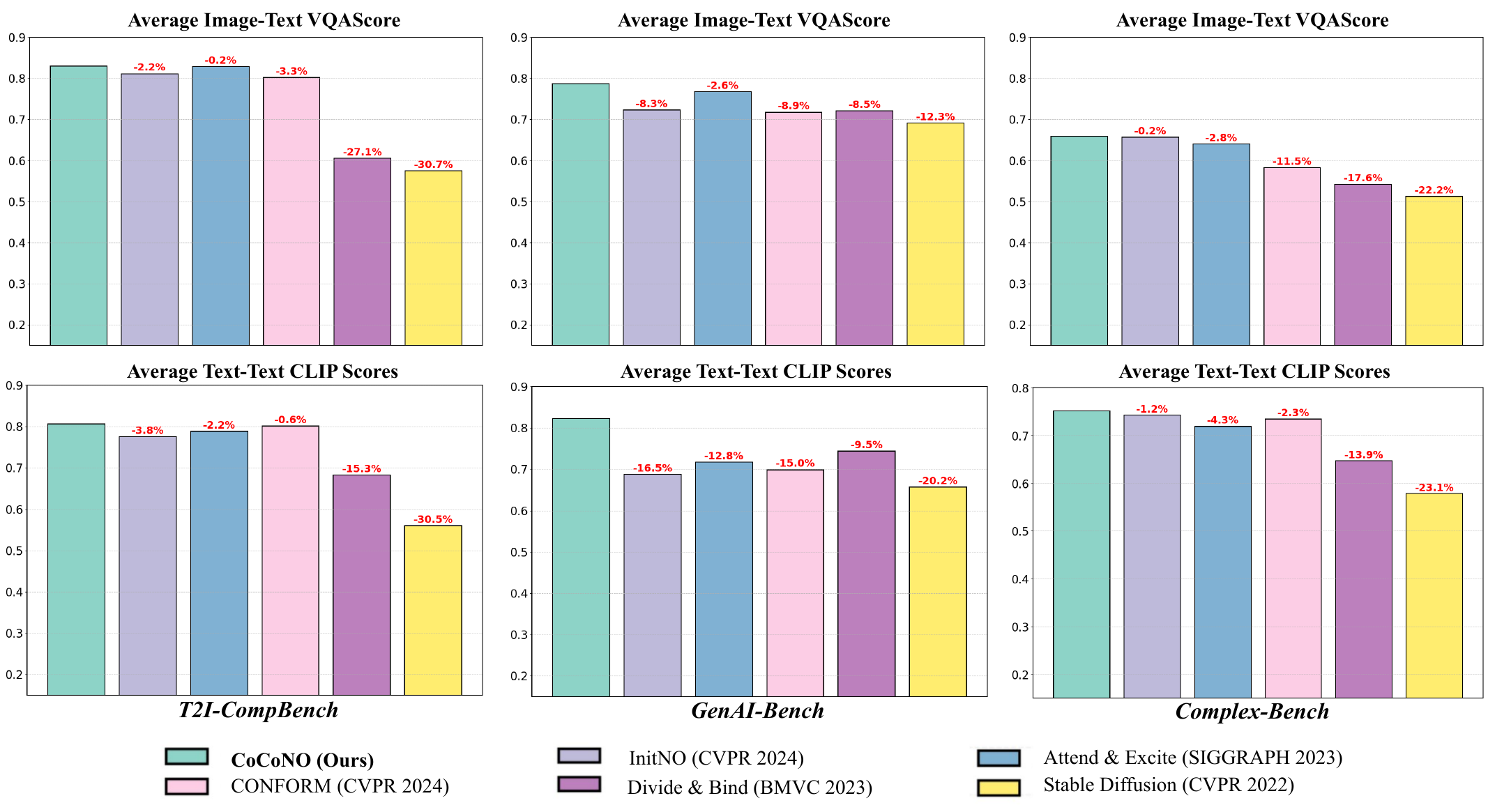}
    \caption{Average image-text and text-text similarities between the text prompts and the images generated by each method.}
    \label{fig:quantGraphs}
\end{figure*}

Following InitNO \cite{guo2024initno}, we also enforce the Kullback-Leibler divergence loss \cite{shlens2014notes, kingma2013auto, van2017neural} to ensure the distribution of the optimized latent remains close to the standard normal distribution: $L_\text{KL}=KL(\mathcal{N}(\mu,\sigma^2\ )||\mathcal{N}(0,1))$

Our overall objective function is then:
\begin{equation}
       L=\lambda_1\ L_\text{Acont}+\lambda_2\ L_\text{Acomp}+\lambda_3\ L_\text{KL} 
\end{equation}
where $\lambda_1=1,\lambda_2=1,\lambda_3=500$ are set empirically.

\section{Experiments}

\subsection{Experimental Settings}

Like InitNO \cite{guo2024initno}, we use Stable Diffusion (SD) v2.1 as our base model. We evaluate our method on the latest benchmarks T2I-CompBench~\cite{huang2023t2i} and GenAI-Bench~\cite{lin2024evaluating}, as well as a set of complex prompts (Complex-Bench) curated with the help of GPT~\cite{brown2020language}. See supplementary for more details. We not only compare to our closest baseline InitNO \cite{guo2024initno}, but also baseline SD and a few other relevant methods: CONFORM~\cite{meral2024conform}, Divide-and-Bind~\cite{li2023divide}, and Attend-and-Excite~\cite{chefer2023attend}.

\begin{figure}[h]
    \centering
    \includegraphics[width = 1.01\linewidth]{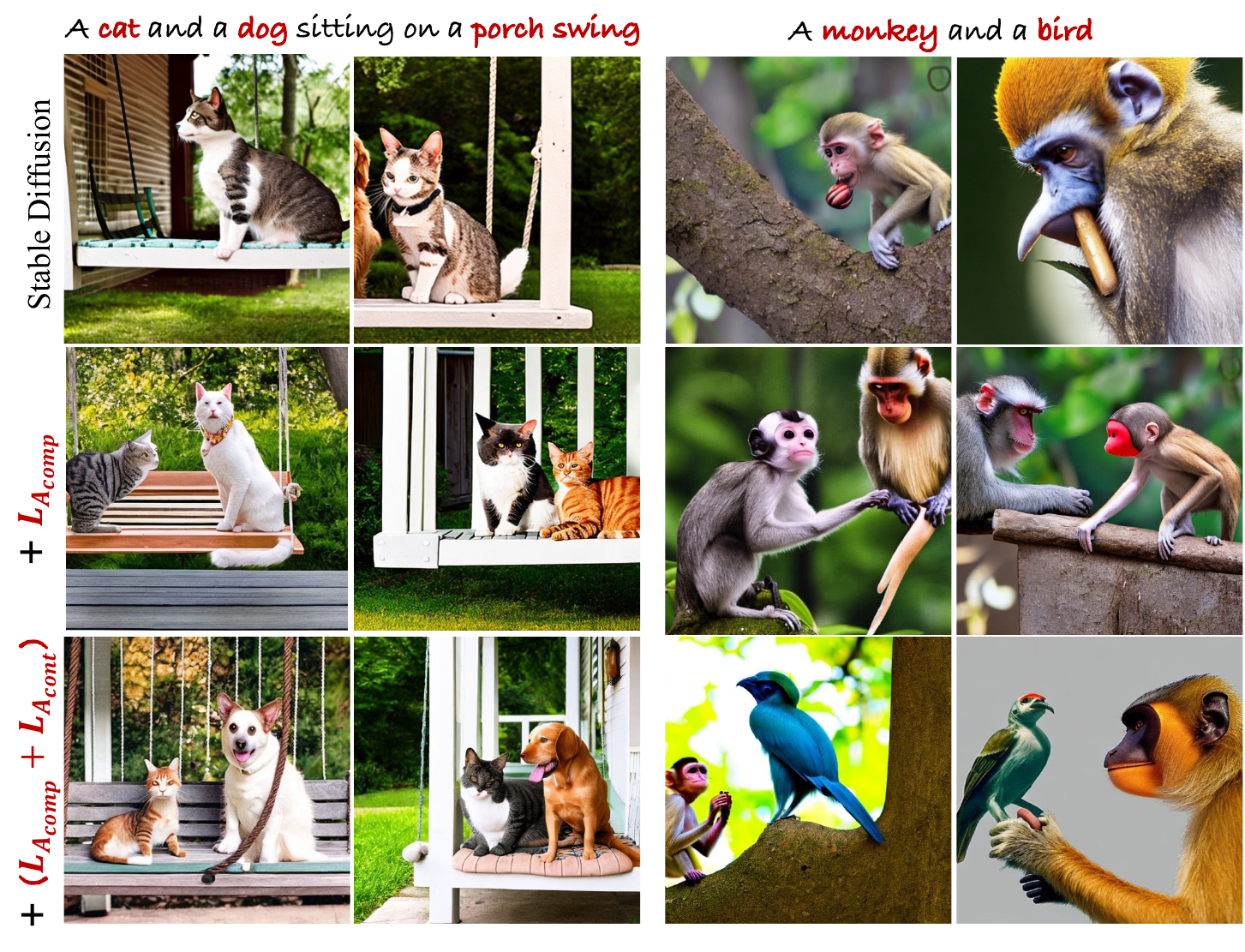}
    \caption{Qualitative ablation results.}
    \label{fig:ablationQual}
\end{figure}

\begin{table}
\centering
\scalebox{1.0}{
\begin{tabular}{@{}l|ccc@{}}
\toprule
Method   & Image-Text  & Text-Text   \\ \midrule
SD \cite{rombach2022high}    & 0.60           & 0.61            \\
SD + $L_{A_{comp}}$  & 0.71 \textcolor{blue}{(+18.3\%)}& 0.74 \textcolor{blue}{(+21.3\%)}   \\
SD + $L_{A_{comp}} + L_{A_{cont}}$ & 0.76 \textcolor{blue}{(+26.7\%)}          & 0.80 \textcolor{blue}{(+31.1\%)}
\\ \bottomrule
\end{tabular}%
}
\caption{Ablation results.}
\label{tab:abl_text_text}
\end{table}

\begin{table}
\centering
\scalebox{1.0}{
\begin{tabular}{@{}c|c@{}}
\toprule
Method                             & Preference \\ \midrule
 Divide \& Bind~\cite{li2023divide} &  $6.9\%$\\
 CONFORM~\cite{meral2024conform} &  $9.1\%$\\
 InitNO~\cite{guo2024initno} &  $11.4\%$\\ \midrule
\textbf{CoCoNO} &  \textbf{72.6}\%\\ \midrule
\end{tabular}
}
\vspace{-8pt}
\caption{User study results.}
\label{tab:userStudy}
\end{table}

\subsection{Results}
\textbf{Qualitative Comparison.} We first present a qualitative comparison with our baselines in Figure~\ref{fig:qualResults} where one can note that CoCoNO clearly outperforms all of them. For instance, in the first column, the issues of subject mixing and neglect are clearly evident across all baselines e.g., the \texttt{parrot} with \texttt{monkey's face} for images generated by Attend-and-Excite~\cite{chefer2023attend}, and missing \texttt{monkey} in case of InitNO~\cite{guo2024initno}. Similarly, from the example shown in second column, one can note that most of these methods generate \texttt{cars} having \texttt{cheetah's spotted texture}. This clearly demonstrates that enforcing non-overlapping attention maps, as used in baselines, is insufficient to tackle subject mixing. Further, observe the results in the fourth column where Attend-and-Excite~\cite{chefer2023attend} and CONFORM~\cite{meral2024conform} have images of \texttt{bears} having \texttt{deer's horns}. Here, InitNO~\cite{guo2024initno} and Divide-and-bind~\cite{li2023divide} have completely neglected generating the deer, whereas CoCoNO correctly captures all the subjects mentioned in the prompt.\\
\textbf{Quantitative Comparison.} We next quantify the improvements with CoCoNO using a recent state-of-the-art text-to-image evaluation metric VQAscore~\cite{lin2024evaluating} as well as text-text similarity scores (between BLIP~\cite{li2022blip} generated captions of images and the corresponding ground truth prompts) following prior works~\cite{chefer2023attend}. For each prompt, we generate 64 images with randomly selected seeds and report results averaged across all generations. As one can note from the graphs in Figure~\ref{fig:quantGraphs}, CoCoNO substantially outperforms all baselines across both metrics and all three benchmarks.\\
\textbf{Ablation Study.} We demonstrate the impact of our proposed losses in Figure~\ref{fig:ablationQual} where we incrementally add them to the baseline Stable Diffusion~\cite{rombach2022high} backbone. For instance, consider the example show in the first column. The images generated by the baseline model, shown in the first row, depict only a single object and fail to include the dog. When our attention complete loss is included in the second row, we see images having two objects since this loss ensures each subject from the prompt has a designated self-attention segment. Finally, after adding the attention contrast loss as well, the generated images not only feature two objects but also ensure that one is a cat and the other a dog. This is because this loss minimises the interference between the cross-attention map of cat (dog) with the self-attention segment of dog (cat). Additionally, we generate $64$ images per prompt for all benchmarks by using one loss at a time and report averaged image-text and text-text similarity scores in Table~\ref{tab:abl_text_text}, where each loss contributes significantly and the best performance obtained when we use both. \\
\textbf{User Study.} Finally, we conduct a user study with the generated images where we show survey respondents a textual prompt, and then ask them to select the images (among sets from four different methods shown in Table~\ref{tab:userStudy}) that best align with the prompt. As shown in Table~\ref{tab:userStudy}, most users prefer images generated by CoCoNO, providing further evidence for the effectiveness of our proposed losses.

\section{Summary}

We considered the problem of initial latent optimization to find a good starting point for subsequent denoising with text-to-image models. We observed that the current state-of-the-art methods in this space, e.g., InitNO \cite{guo2024initno} often produce images that either miss certain subjects or produce images with mixed-up properties of multiple subjects (recall Figure~\ref{fig:teaser_qual}. We identified the underlying causes to be attention interference (cross-attention map of one subject attending to another subject's self-attention segment) and attention neglect (cross-attention maps of multiple subjects attending to the same self-attention segment since a few segments are missing). We proposed two new losses, attention contrast and attention complete, to fix these issues, and demonstrated our method outperforms the current state of the art with extensive evaluations on multiple benchmark datasets.

\begin{appendices}
    
\section{}

We first show more results for both attention interference and neglect issues with baseline InitNO \cite{guo2024initno}, and how these issues get fixed with proposed CoCoNO. This is followed by more implementation details of CoCoNO. In the subsequent sections, we provide additional quantitative results to specifically evaluate the aspect of subject mixing with SAM \cite{kirillov2023segment} and GroundingDINO \cite{liu2023grounding} to estimate number of distinct segments in generated images compared to number of subjects mentioned in the input prompts, and additional qualitative results comparing our method with the pretrained Stable Diffusion \cite{rombach2022high} backbone, and also other relevant methods: InitNO \cite{guo2024initno}, CONFORM~\cite{meral2024conform}, Divide-and-Bind~\cite{li2023divide}, and Attend-and-Excite~\cite{chefer2023attend}. We conclude with some discussions on limitations of our method.

\subsection{Attention Interference and Neglect}
\label{sec:attentionIssues}

As discussed in the main paper, we identified the underlying reasons behind the persistent issues (subject neglect and subject mixing) with existing text-to-image generation frameworks: attention interference and attention neglect. Here, we show more examples demonstrating the same.

\begin{figure*}[]
    \centering
    \includegraphics[width = \linewidth]{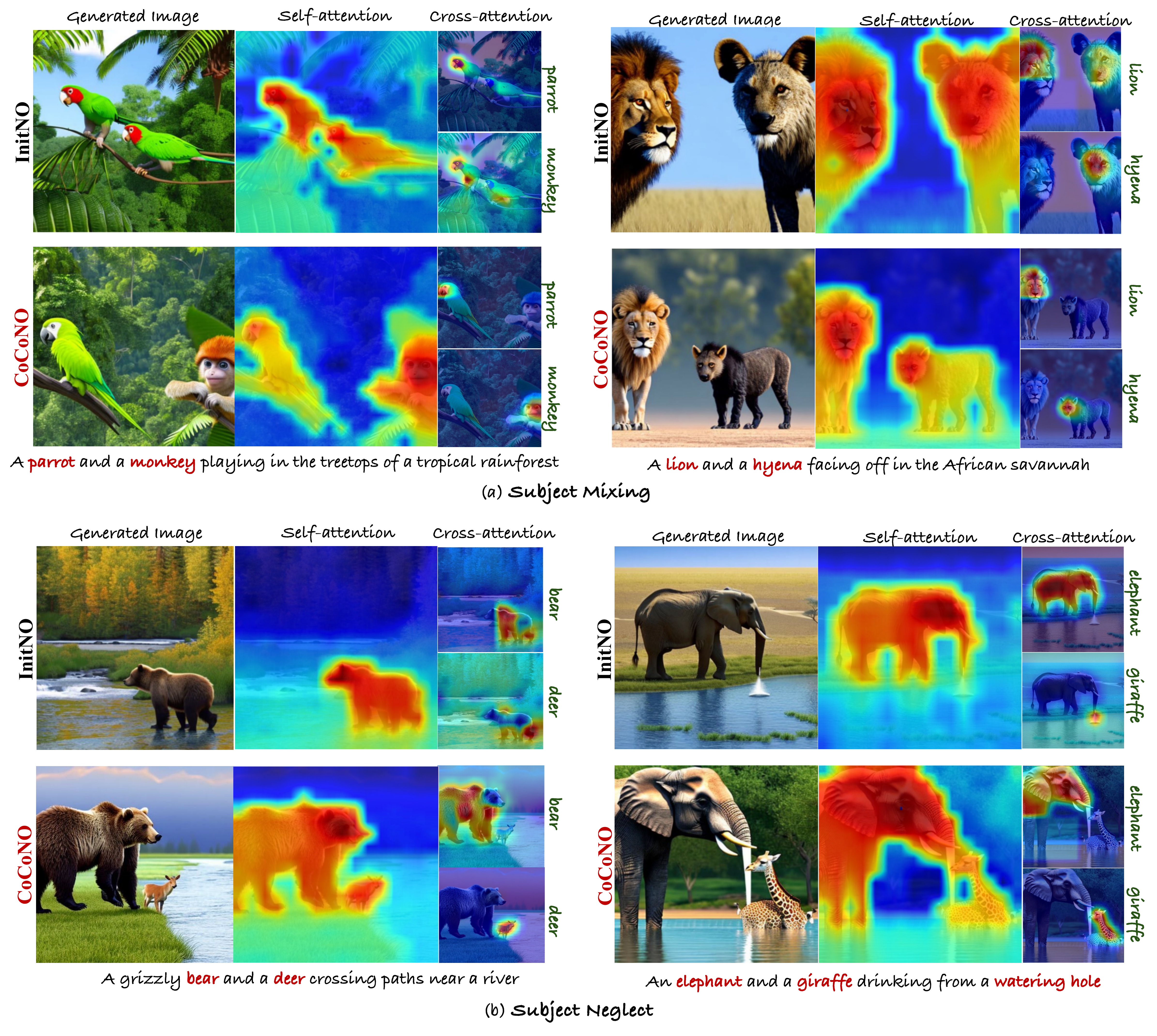}
    \caption{Our proposed CoCoNO alleviates subject neglect and mixing by ensuring one high-response self-attention segment for each subject (e.g., deer and giraffe in second row), while minimizing interference between the cross-attention map for a subject (e.g., monkey in first row) with the self-attention segment of other subjects (parrot here).}
    \label{fig:limitationsSupp}
\end{figure*}

In Figure~\ref{fig:limitationsSupp} (a), we demonstrate the issue of subject mixing (caused by attention interference). We show generated images and the corresponding cross- and self-attention maps using InitNO \cite{guo2024initno} and our proposed method, CoCoNO. Consider the example shown in the first row/first column. Our closest baseline InitNO generates two parrots (instead of a parrot and a monkey), where one can clearly note that the \texttt{parrots} generated have \texttt{monkey-like feet}. This is due to both \texttt{parrot} and \texttt{monkey} attempting to attend to a common object/segment, as can be noted from the self-attention map (notice the high-response regions in \texttt{monkey's} cross-attention maps attend to the segment for \texttt{parrot} in the self-attention map/generated image, and this is exactly same as the regions in \texttt{parrot} having \texttt{monkey's feet}). The same phenomenon can also be observed in the second example in Figure~\ref{fig:limitationsSupp} (a) (\texttt{hyena} has \texttt{lion-like white beard} due to a small patch of lion's cross attention map attending to \texttt{hyena}'s mouth region in self-attention)

In Figure~\ref{fig:limitationsSupp} (b), we show more examples demonstrating the issue of subject neglect (caused by attention neglect). Consider the images generated by InitNO in the example shown in first column/section. Here, both \texttt{bear} and \texttt{deer} have highly activated cross-attention maps, but that does not help as there is only one segment in the self-attention map and both cross-attention regions attend to this segment. Another example is in the second column where, again, since the self-attention map has only one segment for InitNO, cross-attention regions for both \texttt{elephant} and \texttt{giraffe} end up attending to the same segment.

On the other hand, our proposed method CoCoNO fixes the aforementioned issues and generates images including all subjects (e.g. bear and deer, elephant and giraffe in Figure~\ref{fig:limitationsSupp} (b)) without any undesirable inter-object mixing (e.g. parrot and monkey, lion and hyena in Figure~\ref{fig:limitationsSupp} (a)). This further demonstrates the efficacy of our proposed attention completion and contrast losses in tackling the issues highlighted above.

\label{sec:addQual}

\subsection{Implementation Details}
\label{sec:impl}
We use the official Stable Diffusion v2.1 text-to-image model consistently across all results with our proposed method and baselines for a fair comparison. We use $50$ timesteps during the denoising process with a guidance scale of $7.5$ throughout. We applied Otsu thresholding~\cite{yousefi2011image} for binarising the self-attention maps, followed by Breadth First Search~\cite{lee1961algorithm} to obtain the segments. For cross attention maps, we follow the standard gaussian smoothing from prior works \cite{chefer2023attend}. For both cross-attention and self-attention maps, we utilise layers with resolution 16 as they have been shown in prior works \cite{chefer2023attend} to the contain the maximum information. from The subject tokens use during the latent optimization can be specified both manually, or identified automatically using GPT~\cite{brown2020language}.
Additionally, the PCA component computed to obtain the self-attention map could potentially be inverted at times as PCA can yield both positive and negative component weights. To tackle the same, we compute losses using both maps (original as well as inverted) and proceed with the one that gives the minimum loss for further optimization. 

During latent update step, Adam optimizer with a learning rate of $10^{-2}$ is employed to update parameters. Also, some of the starting latents are excessively challenging to optimize given the additional constraint to stay close to $\mathcal{N}(0,1)$. To tackle such scenarios, we impose an upper limit on the number of optimization steps and if the losses have not decreased beyond certain thresholds in a fixed number of steps, we resample $z_T$ and restart the optimization process. 

Lastly, we would like to highlight that our proposed losses are convex within an assignment mapping. An assignment swap while resampling the noise could potentially lead to an increase in loss as well. Therefore, we additionally maintain a cache of the most optimal latent throughout the optimization.

\textbf{Complex-bench. }
We utilised GPT~\cite{brown2020language} to curate a set of $20$ complex prompts. We prompted GPT with the following input query as ``Generate a set of 20 complex prompts in order to create a benchmark for evaluating text-to-image generation models, including some real-world scenarios and objects.". The generated prompts are list below:
\begin{enumerate}
    \item A group of monarch butterflies fluttering around blooming cherry blossom trees during spring.
    \item A steam locomotive crossing an iron bridge over a rushing mountain river in the winter.
    \item A lively farmers' market full of customers buying a variety of fruits and vegetables on a bright sunny day.
    \item A horde of surreal clockwork robots marching through an old-fashioned Victorian-style cityscape at dusk.
    \item A grand medieval castle nestled on a high cliff overlooking a calm sea under a stormy sky.
    \item Firefighters in action, extinguishing a raging forest fire with helicopters and fire trucks.
    \item A pair of wildlife explorers observing a herd of elephants crossing a vast African savannah at sunset.
    \item An extraterrestrial landscape with strange, colorful alien plant life under three moons in the sky
    \item An energetic Mardi Gras parade featuring dozens of vibrant floats and costumed dancers on a bustling street.
    \item Digital nomads working on their laptops at a tranquil beach cafe during a breathtaking sunset.
    \item A giant futuristic city lit up with neon lights and dominated by towering skyscrapers and hovering vehicles.
    \item The idyllic English countryside scene with a traditional thatched-roof cottage surrounded by fields of blooming lavender.
    \item A traditional Japanese tea ceremony taking place in a serene and beautifully landscaped garden in the fall.
    \item An advanced robotics lab filled with engineers and scientists developing humanoid robots.
    \item A children's playground filled with laughter, colourful balloons, and kids playing on a sunny afternoon.
    \item A massive glacier calving into a deep, icy-blue fjord under the northern lights in the Arctic.
    \item A high-stakes poker game occurring in a luxury suite with a view of the Las Vegas Strip at night.
    \item A bustling Moroccan market with stalls selling colourful spices, textiles, and traditional artisan crafts.
    \item A post-apocalyptic city reclaimed by nature, with skyscrapers covered in vines and deer roaming the streets.
    \item An underwater scene of a vibrant coral reef teeming with diverse marine life and a sunken pirate ship in the background.
\end{enumerate}
\textbf{Inference time.} We evaluated and compared the average time taken for image generation ($512$ $\times$ $512$ pixel images) by proposed CoCoNO compared to baseline Stable Diffusion backbone on a single A100 (40GB). We found Stable Diffusion to take $7.71$ seconds while our proposed method takes $17.21$ seconds.

\subsection{Additional Quantitative Results}
\label{sec:addQuant}
\begin{figure*}[]
    \centering
    \includegraphics[width = \linewidth]{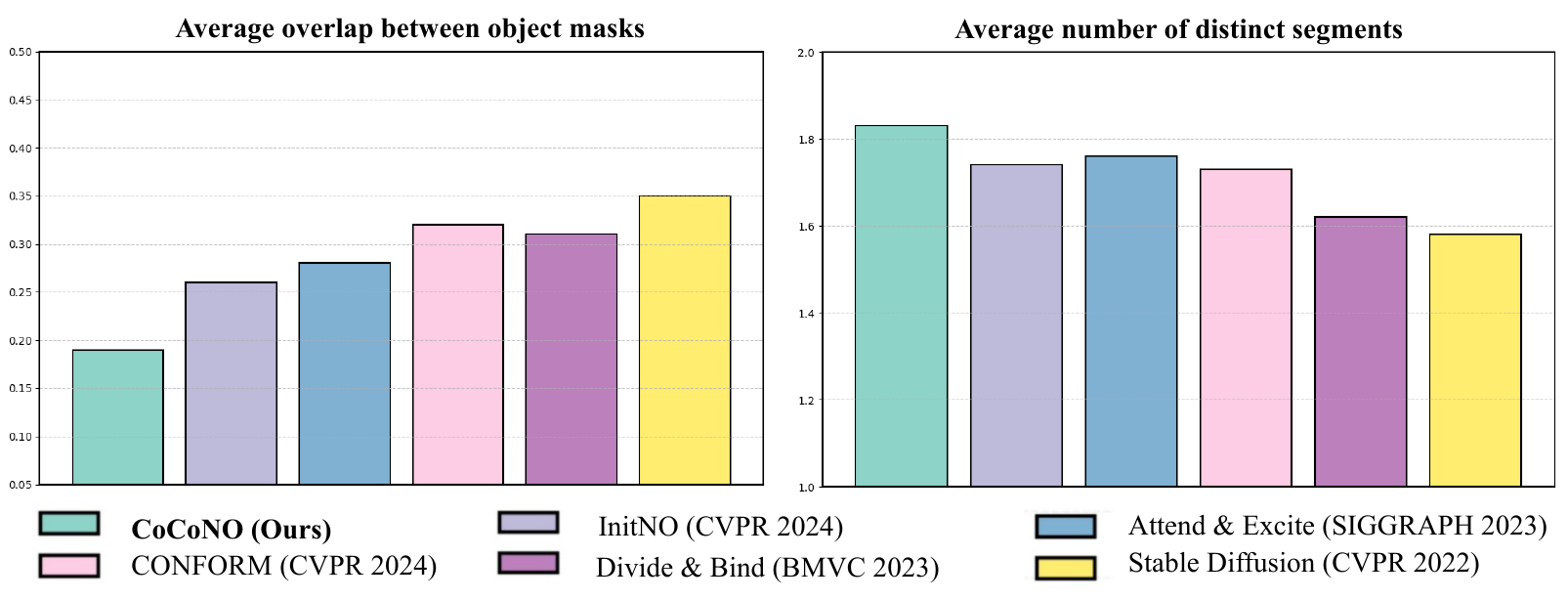}
    \caption{Average number of distinct segments and corresponding overlap score for images generated by each method.}
    \label{fig:quantSupp}
\end{figure*}

In the main paper (Figure 7), we had shown Image-Text and Text-Text similarity comparisons with several existing text-to-image generation methods. Here, in Figure~\ref{fig:quantSupp}, we present additional graphs evaluating the tendencies of subject mixing and neglect in CoCoNO compared to several baselines. We generate a set of 64 images across all three benchmarks using the baselines from the main paper. We then apply Grounded-SAM, a combination of Grounding-DINO \cite{liu2023grounding} and SAM \cite{kirillov2023segment}, to compute segmentation masks for each subject token in the prompt.

Our intuition is that for a prompt containing $n$ subject tokens, the number of distinct segmentation masks obtained via Grounded-SAM, along with the pairwise overlap between these masks, can reveal subject mixing or neglect in the generated images. Specifically:
\begin{itemize}
    \item If the number of distinct masks is less than the number of subjects in the prompt, this indicates subject omission or neglect.
    \item High overlap between masks suggests potential subject mixing, as it means different subjects are associated with the same spatial regions.
\end{itemize}

In Figure~\ref{fig:quantSupp}, we report the average overlap between object masks and the number of distinct segments across all three benchmarks, comparing our CoCoNO approach with relevant baselines. The results show that CoCoNO outperforms others across both metrics: it achieves the lowest overlap, indicating minimal subject mixing, while also producing the most distinct masks, reflecting better subject separation.

\subsection{Additional Qualitative Results}
\begin{figure*}[]
    \centering
    \includegraphics[width = \linewidth]{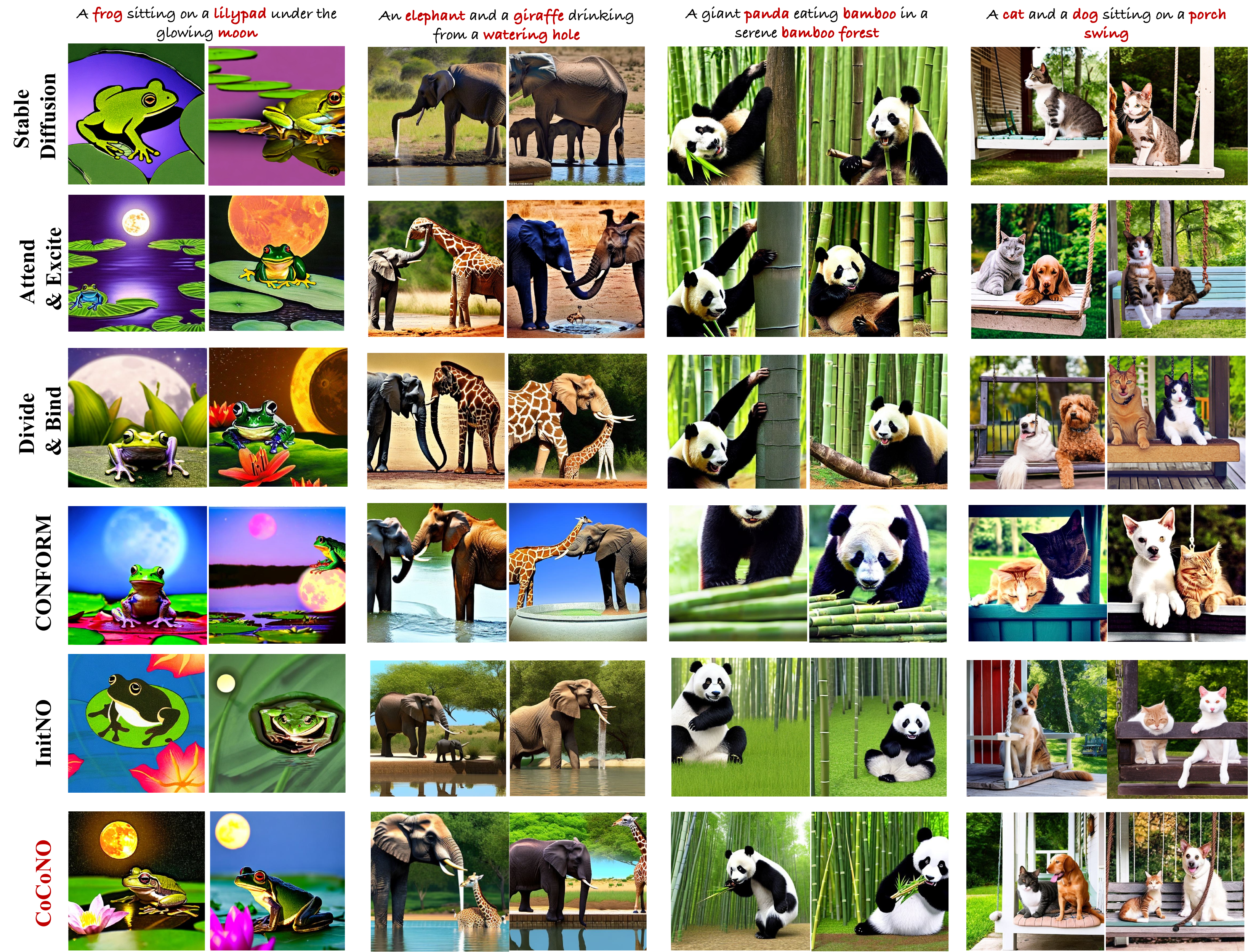}
    \caption{Additional Qualitative comparisons of CoCoNO with recent state-of-the-art methods.}
    \label{fig:qualResultsSupp1}
\end{figure*}

\begin{figure*}[]
    \centering
    \includegraphics[width = \linewidth]{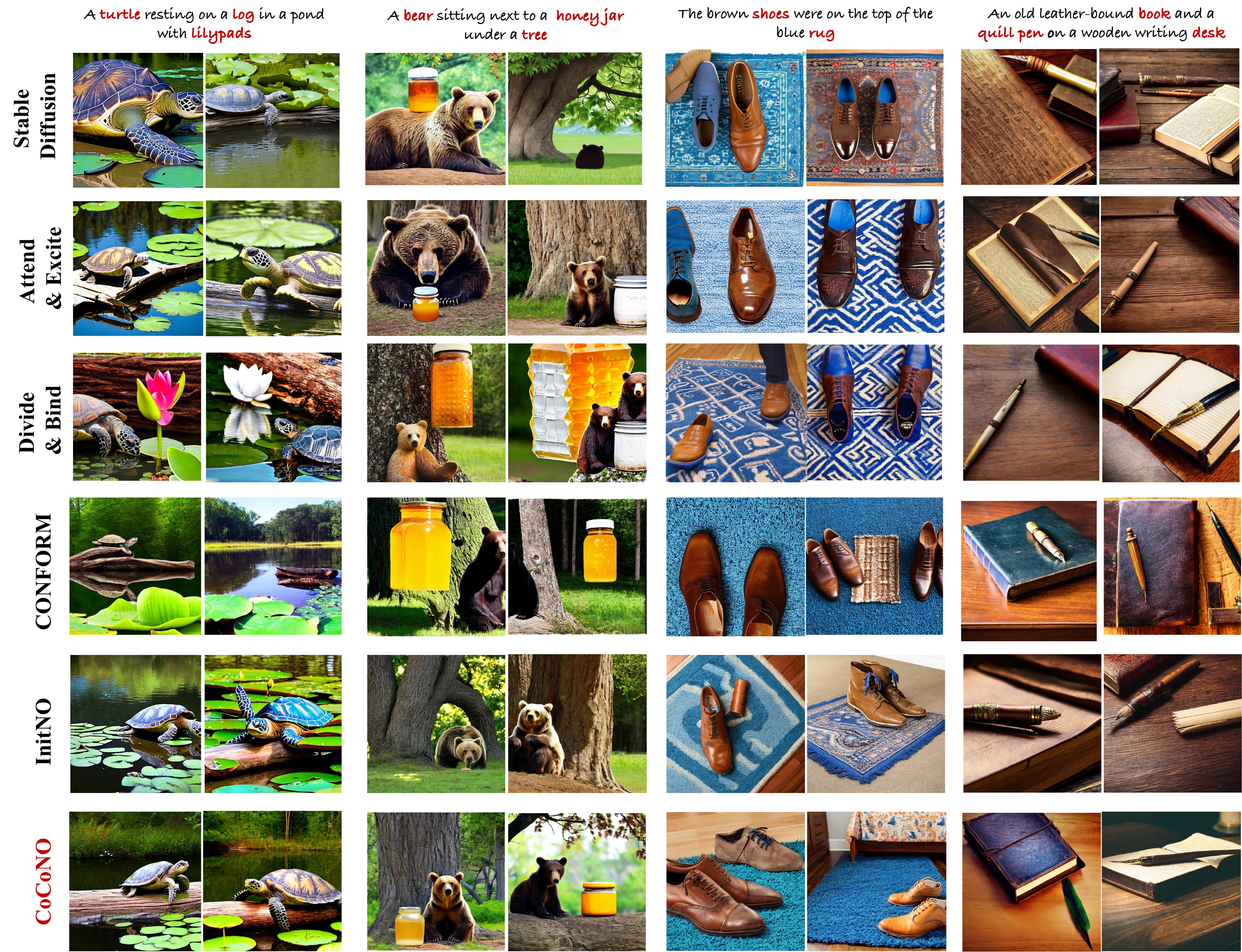}
    \caption{Additional Qualitative comparisons of CoCoNO with recent state-of-the-art methods.}
    \label{fig:qualResultsSupp2}
\end{figure*}

In Figure~\ref{fig:qualResultsSupp1} and ~\ref{fig:qualResultsSupp2}, we show additional qualitative results comparing the performance of our proposed method with pretrained Stable Diffusion \cite{rombach2022high} backbone, and also other relevant methods: InitNO \cite{guo2024initno}, CONFORM~\cite{meral2024conform}, Divide-and-Bind~\cite{li2023divide}, and Attend-and-Excite~\cite{chefer2023attend}. In each case, CoCoNO clearly outperforms all the baselines. For instance, notice the elephant with giraffe's texture when using Divide-and-Bind in second column in Figure~\ref{fig:qualResultsSupp1}, and the shoes having a mix of brown and blue in the third column in Figure~\ref{fig:qualResultsSupp2} across majority baselines. Similarly, one can also observe instances of omitted objects e.g. the giraffe and the honey jar/bear in second columns of Figure~\ref{fig:qualResultsSupp1} and ~\ref{fig:qualResultsSupp2} respectively, whereas CoCoNO correctly captures all the subjects mentioned in the prompt.
\label{sec:addQual}

\subsection{Limitations}
\label{sec:limitations}
CoCoNO is constrained by the backbone Stable Diffusion model's capabilities in modelling inter-subject relations as we currently do not explicitly model the relationships between subjects. That said, if a computational model that captures these relationships were available, it could potentially be integrated into our losses to reflect these relations in the final output.

Additionally, since some starting latents are challenging to optimize given the constraint to stay close to $\mathcal{N}(0,1)$, the optimization process ends up taking considerably high time in reaching a good latent in such scenarios. 

\end{appendices}

{
    \small
    \bibliographystyle{ieeenat_fullname}
    \bibliography{main}
}

\end{document}